%% file: main.tex
\documentclass[10pt,twocolumn,letterpaper]{article}

\usepackage[pagenumbers]{cvpr} 

\input{preamble}

\usepackage[utf8]{inputenc} 
\usepackage[T1]{fontenc}    
\usepackage{multirow}
\definecolor{cvprblue}{rgb}{0.21,0.49,0.74}
\usepackage[pagebackref,breaklinks,colorlinks,citecolor=cvprblue]{hyperref}

\title{`Eyes of a Hawk and Ears of a Fox': Part Prototype Network for Generalized Zero-Shot Learning}

\author{Joshua Feinglass$^1$ \quad Jayaraman J. Thiagarajan$^2$ \quad Rushil Anirudh$^2$ \quad T.S. Jayram$^2$ \quad Yezhou Yang$^1$\\
\textsuperscript{1} Arizona State University\\
\textsuperscript{2} Lawrence Livermore National Lab\\
{\tt\small \{joshua.feinglass,yz.yang\}@asu.edu, \{jjayaram,anirudh1,thathachar1\}@llnl.gov}
}

\begin{document}
\maketitle
\input{sec/0_abstract}    
\input{sec/1_intro}
\input{sec/2_related_work}
\input{sec/3_preliminaries}
\input{sec/4_vinvl}
\input{sec/5_ppn}
\input{sec/6_experiments}

\input{sec/7_conclusion}

{
    \small
    \bibliographystyle{ieeenat_fullname}
    \bibliography{main}
}


\end{document}

%% file: preamble.tex
\usepackage[dvipsnames]{xcolor}


%% file: sec/0_abstract.tex
\begin{abstract}
Many approaches in Generalized Zero-Shot Learning (GZSL) are built upon base models which consider only a single class attribute vector representation over the entire image. This is an oversimplification of the process of novel category recognition, where different regions of the image may have properties from different seen classes and thus have different predominant attributes. With this in mind, we take a fundamentally different approach: a pre-trained Vision-Language detector (VINVL) sensitive to attribute information is employed to efficiently obtain region features. A learned function maps the region features to region-specific attribute attention used to construct class part prototypes. We conduct experiments on a popular GZSL benchmark consisting of the CUB, SUN, and AWA2 datasets where our proposed Part Prototype Network (PPN) achieves promising results when compared with other popular base models. Corresponding ablation studies and analysis show that our approach is highly practical and has a distinct advantage over global attribute attention when localized proposals are available. 
\end{abstract}

%% file: sec/1_intro.tex
\section{Introduction}
Generalized Zero-Shot Learning (GZSL) has become a popular research topic with a wide variety of different approaches \cite{gzslreview2022}. As benchmarks have become more competitive, many researchers have begun to develop increasingly sophisticated models requiring a large amount hyperparameter tuning and multiple stages of computationally expensive training. This hinders progress and introduces a need for potentially prohibitively expensive model reconfiguration and computational overhead in a field originally motivated by the expense of human annotation. As such, we explore the potential of using a general pre-trained Vision-Language (VL) detector model combined with a specialized base model that can perform well out-of-the-box and be trained in a single stage to potentially serve as an improved foundation for more sophisticated enhancements like generative models \cite{tfvaegan2020} and graph networks \cite{rgen2020}.
\begin{figure}
\centering
\def\svgwidth{\columnwidth}
\includegraphics[width=\columnwidth]{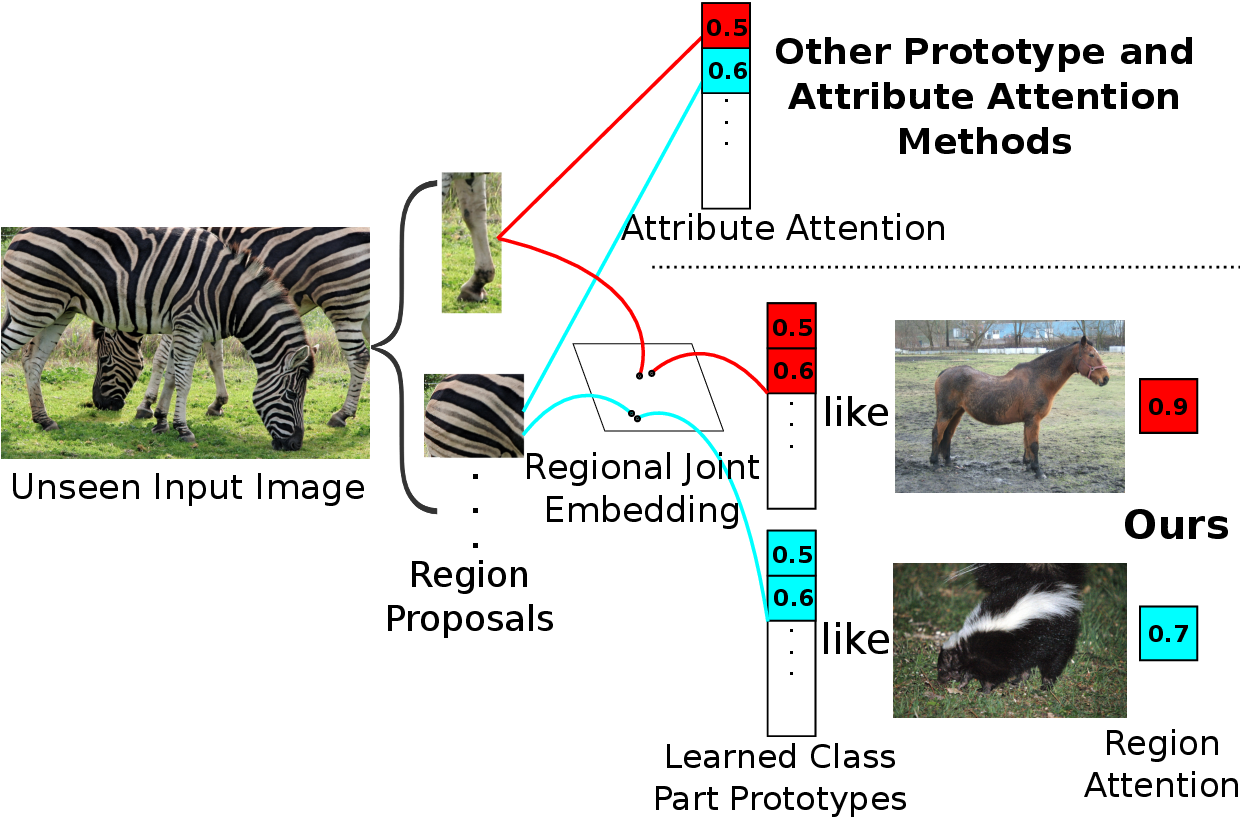}
\caption{A comparison between the proposed Prototype Proposal Network (PPN) approach and existing approaches which utilize global attribute attention like the base model DAZLE \cite{dazle2020}.}
\label{fig:main}
\end{figure}
\par
Localization has been shown to be a key step in many Vision-Language (VL) tasks, especially detail-oriented tasks like fine-grained Zero-Shot Learning \cite{elope2020,proto2020,dazle2020,dazlenips2020,vinvl2021,gaze2021}. In current approaches, all attribute-specific localization is performed by the GZSL model after either extracting learned features or utilizing popular visual features pre-trained using image classification on ImageNet \cite{ugly2018}. To perform this localization, GZL models must correlate global representations of attributes and regions. However, we argue that localized and attribute-specific features can be obtained from VL pretrained detectors like VINVL and GZSL performance can be improved by instead constructing region-specific attribute representations using part prototypes. This is a natural extension of two commonly used base models: SJE \cite{sje2015} which utilizes global joint embeddings of visual features and attributes and DAZLE \cite{dazle2020} which employs visual and attribute attention to generate global representations compared with similarity scoring. These two base models are selected for comparison and enhancement due to their simplicity in terms of both computational complexity and hyperparameter tuning. Further motivation for part-based representations can be found in recent GZSL works which seek to learn more discriminative attribute localizations \cite{dppn2021} based on part structures \cite{domain-invariant2023}. Figure \ref{fig:main} shows an example where a model must generalize to the unseen class, zebra. While prior approaches are limited to estimating the likelihood that an attribute (e.g. stripes) is present in the entire image, our approach can deduce that the shape of a zebra's legs resemble those of a horse, while the color and patterns of its hair is similar to that of a skunk. 
\par
We evaluate our approach using the popular GZSL benchmark provided by \cite{ugly2018} which utilizes the AWA2 \cite{ugly2018}, CUB \cite{cub2010}, and SUN \cite{sun2012} datasets. Ablation studies of different region localization sources, two proposed regularizers, and a proposed post-processing based calibration technique are performed in support of our framework. We observe that the performance of PPN greatly improves when utilizing pre-trained visual feature extractors with more localized information, resulting in very promising performance against comparable base models.
\noindent \textbf{Contributions.}
We demonstrate the potential using a pre-trained VL detector for GZSL and propose a novel base model, PPN, designed to better leverage the localized region proposals to achieve promising results on popular GZSL benchmarks.

%% file: sec/2_related_work.tex
\section{Related Work}
\textbf{Pre-Trained Localization} with object detectors for GZSL has been previously been attempted with dataset-specific detectors \cite{partdet2020}, yielding sub-optimal results when compared against dataset-specific attention mechanisms \cite{proto2020,dazle2020}. This performance gap when using an object detector may be due to a reliance on dataset-specific part annotations, which is both impractical and biased compared to general vision-language detector pre-training. Furthermore, pre-trained Vision-Language models trained to provide global representations \cite{vit2021} have shown promising results in certain GZSL settings \cite{i2mv2022,i2d2022} when combined with sophisticated transformer-based architectures and retrieved class information. While there is evidence to suggest that the performance of pre-trained localization methods may suffer when applied to unaligned tasks \cite{misalignment2024,mess2023}, pre-trained models have shown surprising resilience in generalization tasks like zero-shot classification \cite{clip2021} and out-of-distribution detection \cite{ood2023,covariate2022}. We postulate that the object and attribute pre-training utilized by VINVL \cite{vinvl2021} can potentially serve as a strong foundation for localization in GZSL and allow for methods to explore improvements in other components of the inference process, like attribute representation.\\
\textbf{Specialized GZSL Localization} is usually a computationally expensive process for competitive methods \cite{review2022}, with popular approaches often utilizing Generative Adversarial Networks \cite{gan2018,tfvaegan2020,fvaegan2019} and Variational Auto-Encoders \cite{vae2019,vae2022,vae2020}. These methods require dataset-specific effort like precise hyperparameter tuning and multiple-stages of training. \\ 
\textbf{Structural Misalignment} between seen and unseen data continues to be a major bottleneck for GZSL performance \cite{structaligned2021}. Prior works have attempted to address these issues with image domain transformations \cite{hierachy2021,structaligned2021}, part-object relations \cite{partobject2024}, and attribute attention \cite{dazle2020,dazlenips2020}. Our work further extends attribute attention by associating attributes with part prototypes which aim to describe the typical \cite{typicality1997,smurf2021} characteristics of a class and can potentially serve as more robust primitives than less localized image information.


%% file: sec/3_preliminaries.tex
\begin{figure*}[!t]
\centering
\def\svgwidth{\columnwidth}
\includegraphics[width=\textwidth,height=\textheight,keepaspectratio]{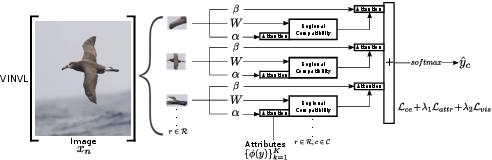}
\caption{A visualization of the proposed Part Proposal Network (PPN) methodology. $\alpha$, $W$, and $\beta$ represent learned parameters and correspond to the part prototypes, regional embedding, and the mapping function for regional attention, respectively.}
\label{fig:method}
\end{figure*}
\section{Preliminaries}
\subsection{(Generalized) Zero-Shot Learning}
The (Generalized) Zero-Shot Learning (GZSL) task evaluates model performance under label shift. The training set is defined as
$\{(x_n,y_n)|x_n \in X^s,y_n \in Y^s\}^{N_s}_{n=1}$ where images $x_n$ and their corresponding labels $y_n$ are sampled from seen classes $Y^s$ exclusively. In the Zero-Shot Learning (ZSL) setting, models predict on new examples such that
$\mathcal{X} \rightarrow \mathcal{Y}^u$ where $\mathcal{Y}^u$ refers to unseen classes not used for training. In the generalized setting, models predict on examples from both the seen and unseen classes
$\mathcal{X} \rightarrow \mathcal{Y}^u \cup \mathcal{Y}^s$. ZSL differs from other label shift generalization benchmarks in that for each class label $y\in \mathcal{Y}^u \cup \mathcal{Y}^s$, attribute information $\phi(y)$ is available for use at both train and test time.
\subsection{Existing Approaches in GZSL}
For classification, prior Zero-Shot Learning methods rely on a compatibility function $\psi_c$ \cite{ale2013,sje2015} between image features $\theta(x)$ and class attribute vectors $\phi_c(y)$ to estimate the likelihood of $y$ being class $c$ for $c \in \mathcal{C}$
\begin{equation}
P(y=c)=\psi_c(x,y) = \theta(x) \otimes \phi(y_c).
\end{equation}
Regularization of the image feature representation has been the primary focus of GZSL research \cite{fvaegan2019,tfvaegan2020}, while more recent methods have begun to apply grid cell \cite{dazle2020} attention and pixel-level \cite{proto2020} attention to the image representation $\Theta(x) \rightarrow \Theta_\mathcal{R}(x)$.
\subsection{Vision-Language Detectors}
Vision-Language (VL) Detectors \cite{butd2018,vinvl2021} differ from general object detectors in that they are designed to provide detailed labels and features for both objects and their corresponding attributes. This makes them uniquely suited for common Vision-Language tasks like image captioning, VQA, and text-to-image/image-to-text retrieval. The VINVL \cite{vinvl2021} detector is extensively pre-trained across Open Images \cite{openimages2020}, Objects375 \cite{obj3652019}, COCO \cite{coco2014}, Visual Genome \cite{vg2016} with standard object annotations before being fine-tuned on Visual Genome with both an object detection and attribute classification loss has demonstrated State-of-the-Art results across a wide variety of Vision-Language tasks. The region-specific features provided by VINVL set serve as a flexible image-region feature extraction method, providing regional features $\theta_r$ for all proposed regions of interest, $r=\{1,...,R\}$, as shown for an image $x$
\begin{equation}
    \{\theta_r\}_{r=1}^R = \text{VINVL}(x).
\end{equation}

%% file: sec/4_vinvl.tex
\section{VL Detector based ZSL Architectures}
Encoder-decoder architectures, where a general encoder is trained across a large and diverse set of data and a specialized decoder is fine-tuned on a small dataset, dominate the current Machine Learning landscape. We theorize that a similar paradigm should be employed to ZSL, where data scarcity makes fine-grained classification tasks exceptionally challenging. To this end, our work establishes that VL detectors can serve as an encoder for the ZSL tasks, effectively replacing the visual attention and feature regularization approaches of prior works with object and attribute localization and classification for pre-training. Thus, our work places greater emphasis on compatible decoders to create modular ZSL architectures. 
\par
With its use of grid-cells as input, DAZLE \cite{dazle2020} serves as the only VINVL compatible decoder in existing ZSL architectures. In an ablation study in Section 
\ref{sec:gzsl}, we show that simply substituting the ImageNet classify grid-cell features with the regional features provided by VINVL provides a significant boost to the achievable performance of DAZLE. However, the immediate aggregation of VINVL region proposals performed by DAZLE in this architecture removes the regional information provided by VINVL before measuring the compatibility between visual and attribute features. Thus, we explore the use of regional information in compatibility measures with our proposed network.

%% file: sec/5_ppn.tex
\setlength{\tabcolsep}{4pt}
\renewcommand{\arraystretch}{1.2} 
\begin{table*}[t] 
\centering
 \resizebox{.85\linewidth}{!}{%
   \begin{tabular}{l | l | c c c |c c c |c c c |c c c }
   & & \multicolumn{3}{c|}{\textbf{Zero-Shot Learning}} & \multicolumn{9}{c}{\textbf{Generalized Zero-Shot Learning}} \\
  	\textbf{Visual} & & \textbf{AWA2} & \textbf{CUB} & \textbf{SUN} & \multicolumn{3}{c}{\textbf{AWA2}} & \multicolumn{3}{c}{\textbf{CUB}} & \multicolumn{3}{c}{\textbf{SUN}}  \\
  \textbf{Features}\ \ \ \ \ & \textbf{ZSL Model} & \textbf{T1} & \textbf{T1} & \textbf{T1}& \textbf{u} & \textbf{s} & \textbf{H} & \textbf{u} & \textbf{s} & \textbf{H} & \textbf{u} & \textbf{s} & \textbf{H}  \\
  	
  	\hline
  	
  	\multirow{2}{2em}{ResNet101 \cite{resnet2016}} 
    & SJE~\cite{sje2015} & 61.9 & 53.9 & 52.7 & 8.0 & 73.9 & 14.4 & 23.5 & 59.2 & 33.6 & 14.4 & 29.7 & 19.4\\
    & DAZLE~\cite{dazle2020}
  	& 66.2 & 66.0 & 60.3 & 57.5 & 76.2 & \textbf{\textcolor{red}{65.5}} & 56.8 & 59.7 & 58.2 & 48.4 & 26.4 & 34.1 \\
    & PPN (Ours) & 58.6 & 55.8 & 54.1 & 51.4 & 70.4 & 59.4 & 46.2 & 50.5 & 48.3 & 21.7 & 24.8 & 23.2 \\
  		\hline
  	\multirow{2}{2em}{APN(feats) \cite{proto2020}} 
     & SJE \cite{sje2015} (APN-base)
  	& \textbf{\textcolor{red}{68.4}} & 72.0 & 61.6 & 56.5 & 78.0 & \textbf{\textcolor{red}{65.5}} & 65.3 & 69.3 & \textbf{\textcolor{blue}{67.2}} & 41.9 & 34.0 & 37.6 \\
    & DAZLE~\cite{dazle2020}
  	& - & 52.9 & - & - & - & - & 42.7 & 57.8 & 49.1 & - & - & - \\
    & PPN (Ours)& - & 64.8 & - & - & - & - & 50.2 & 66.8 & 57.3 & - & - & - \\
  		\hline
	\multirow{2}{2em}{VINVL \cite{vinvl2021}} & DAZLE~\cite{dazle2020}
  	& 63.2 & \textbf{\textcolor{red}{74.8}} & \textbf{\textcolor{blue}{66.0}} & 54.7 & 70.4 & 61.6 & 66.9 & 63.1 & 64.9 & 53.8 & 33.1 & \textbf{\textcolor{blue}{41.0}} \\
  	& PPN (Ours) $base$
  	& 54.7 & 64.8 & 62.9 & 30.4 & 85.8 & 44.9 & 50.2 & 66.8 & 57.3 & 45.2 & 34.7 & 39.3 \\
   & \textbf{\rotatebox[origin=c]{180}{$\Lsh$}} $\mathcal{L}_{vis}$
    & 63.9 & 65.9 & 65.1 & 40.3 & 82.6 & 54.2 & 54.9 & 65.5 & 59.7 & 50.6 & 33.5 & 40.3 \\
    & \textbf{\rotatebox[origin=c]{180}{$\Lsh$}} $\mathcal{L}_{attr}$
    & 70.4 & 72.1 & 63.5 & 60.1 & 62.8 & 61.4 & 61.9 & 65.0 & 63.4 & 48.6 & 31.2 & 38.0 \\
    & \textbf{\rotatebox[origin=c]{180}{$\Lsh$}} $\mathcal{L}_{attr}\!+\!\mathcal{L}_{vis}$
  	& \textbf{\textcolor{blue}{70.4}} & \textbf{\textcolor{blue}{76.0}} & \textbf{\textcolor{red}{65.0}} & 59.2 & 75.9 & \textbf{\textcolor{blue}{66.6}} & 65.8 & 67.8 & \textbf{\textcolor{red}{66.8}} & 48.6 & 32.5 & \textbf{\textcolor{red}{39.0}} \\
  \end{tabular}
}
\small
\caption{A comparison of the performance of PPN and popular base models when utlizing different visual feature extractors in the human-annotated attribute ZSL and GZSL setting with the proposed split from \cite{ugly2018}. For GZL, accuracy per unseen class (\textbf{u}), accuracy per seen class (\textbf{s}), and their harmonic mean (\textbf{H}) are all reported. Visual feature extractors lower in the table provide more localized feature information. Parameters are set at $\lambda_{1}=0.1$, $\lambda_{2}=0.1$, and $z=10^8$ (multiplicative calibrated stacking) for all PPN variants. Word2vec embeddings \cite{w2v2013} are used for the attribute representations of all reported results to ensure a fair comparison. The best results are highlighted in \textbf{\textcolor{blue}{blue}} (best) and \textbf{\textcolor{red}{red}} (second best) for each evaluation with the combination of VINVL and PPN utilizing both regularizers consistently achieving either the best or second best results when compared with other approaches.}
\label{tab:ZSL_acc}
\end{table*}
\section{Part Prototype Network}
\subsection{Proposed Architecture}
Priors of each attribute for a given class $\phi^a$ in space $\mathcal{R}^{C \times A}$ are provided by human annotation by ZSL datasets where $a \in A$ represent the human selected attributes. Using the names provides for each attribute, a semantic representation of embedding length $K$ can be constructed for each attribute $\phi_k$ in $\mathcal{R}^{A \times K}$ using word2vec \cite{dazle2020,w2v2013}. We proceed to combine these two information sources by expanding the attribute priors along the semantic embedding dimension and the semantic embeddings of the attributes along the class dimension and performing a hadamard product which yields a semantic class attribute tensor $\phi^{a}_k$ in the space $\mathcal{R}^{C \times A \times K}$. 
\par
Visual features are provided by VINVL \cite{vinvl2021} for each proposed region of interest. These are used for both region-specific class semantic representation $\textbf{f}_c^r$ extracted from the semantic class attribute tensor using attribute attention and a visual semantic embedding which is compared against the corresponding class semantic embedding for that region as shown in Figure \ref{fig:method}.
\begin{equation}
\textbf{f}_c^r = g(x,y_c)=\sum_{a=1}^A[\alpha(\theta_r(x))]_a \times \phi(y_c)^a_k,
\end{equation}
where a linear combination in the word2vec semantic space of all $A$ attribute embeddings is taken for each class $c \in \mathcal{C}$ using learned class part prototype $\alpha$. $\alpha$ serves as a part-specific extension from the global attribute attention utilized in DAZLE \cite{dazle2020} and is intended to extract attribute information relevant to the parts present in the input region. Note that $\alpha$ shares the same parameters for all attention mappings. Extending from the global compatibility functions proposed in prior ZSL works \cite{ale2013,sje2015}, we aggregate the compatibility computed for each region and class into an overall compatibility function for each class as shown
\par
\begin{equation}
\psi_c(x,y_c) = \sum_{r=1}^ R[\theta_r(x)]^T \textbf{W} \textbf{f}_c^r\beta(\theta_r(x)),
\end{equation}
where $\textbf{W}$ is the trained regional embedding which maps the region proposals from VINVL into the word2vec semantic space and $\beta$ is a learned function which maps a region proposal to the attention for its compatibility function. Like $\alpha$, parameters of function $W$ and $\beta$ are also shared across all regions such that a universal mapping between the image and semantic space is learned. Applying a softmax across each compatibility function of each class
\begin{equation}
\hat{y}_c=\frac{\text{exp}\{\psi_c(x,y_c)\}}{\sum_i\text{exp}\{\psi_c(x,y_i)\}},
\end{equation}
results in probability mass function $\hat{y}$ which estimates the likelihood that the image example belongs to any given class.
\subsection{Loss and Regularization}
The cross-entropy between the probability mass functions of our model's prediction and the one-hot ground-truth class label
\begin{equation}
\mathcal{L}_{ce}(\{\hat{y}_c\}_{c \in \mathcal{C}})=-\sum_c y_c\text{log}(\hat{y}_c),
\end{equation}
serves as the primary task for our optimization.
\par
For our regularization terms, we construct a penalty function to ensure our attribute and visual representations of the image are relevant to unseen class attributes based on the average of the unseen attribute priors $\phi^a(y \in \mathcal{Y}^u)$ provided by human annotation
\begin{equation}
H(\phi^a) = 1-\frac{1}{|\mathcal{Y}^u|}\sum_{y \in \mathcal{Y}^u} \phi^a(y).
\end{equation}
Penalties for each attribute vary in the range of 0 to 1, with a lower value indicating that the attribute in question occurs more frequently in unseen classes. For attributes, this penalty is multiplied by the square of the attribute attention. Squaring the attention weight ensures outliers incur a greater penalty. This scaled penalty is then summed over all attributes and averaged over all regions as shown
\begin{equation}
\mathcal{L}_{attr}(\phi^a,x)=\frac{1}{R}\sum_{r=1}^R \sum_{a=1}^A H(\phi^a)[\alpha(\theta_r(x))]_a^2.
\end{equation}
For visual semantic features, the penalty is projected into the w2v space using the w2v embeddings of the attribute classes $\phi_k$ and a cosine embedding loss is used to contrast the attention-weighted aggregation of the VINVL region proposals projected into the word2vec space $\sum_{r=1}^R[\theta_r(x)]^T\textbf{W}\beta(\theta_r(x))$ with the penalty with the projected penalty
\begin{equation}
\mathcal{L}_{vis}=max(0,cos(\sum_{r=1}^R[\theta_r(x)]^T\textbf{W}\beta(\theta_r(x)),\sum_{a=1}^A H(\phi^a)\phi_k)).
\end{equation}
\par
The cross-entropy, prior confidence, and self-calibration loss functions are combined to form the training objective
\begin{equation}
\min_{\textbf{W},\alpha, \beta}\  \mathcal{L}_{ce}+\lambda_1\mathcal{L}_{attr}+\lambda_2\mathcal{L}_{vis},
\end{equation}
which optimizes the joint embedding and attention parameters.
\subsection{Pre/Post-Processing}
$L_{2}$ normalization is performed across the attribute dimension $A$ for our attribute input tensor $\mathcal{R}^{C \times A \times K}$ as described in \cite{preprocess2021} and the feature dimension for region proposals. Because ZSL models are only exposed to seen classes at training time, their confidence is typically biased towards predicting seen classes. Calibrated stacking is a standard post-processing approach for adjusting confidence bias of seen classes \cite{proto2020}. Since part prototypes in PPN are constructed based on seen classes, the prediction confidence of PPN is significantly higher for seen classes, even when compared against prior methods. For example, APN \cite{proto2020} report using additive calibrated stacking value of 0.8 for CUB while our method would require a value of 0.9995 to achieve its best validation set performance. To address this phenomenon, we propose multiplicative calibrated stacking as a means of adjusting confidence bias where prediction confidences corresponding to a seen class are adjusted by dividing by a constant $z$ as shown
\begin{equation}
\hat{y}
    = \begin{cases} 
      \frac{\hat{y}_c}{z}\ \ \  if\  y \in \mathcal{Y}^s, \\
      \hat{y}_c\ \ \ otherwise. \\
   \end{cases}
\end{equation}
With addition, all predictions experience the same adjustment while multiplication applies less and potentially no adjustment to predictions with little or no confidence, respectively. This allows us to apply more significant confidence adjustments without impacting false positive rate of unseen selections as significantly.

%% file: sec/6_experiments.tex
\begin{figure}
\centering
\def\svgwidth{\columnwidth}
\includegraphics[width=\columnwidth,height=\textheight,keepaspectratio]{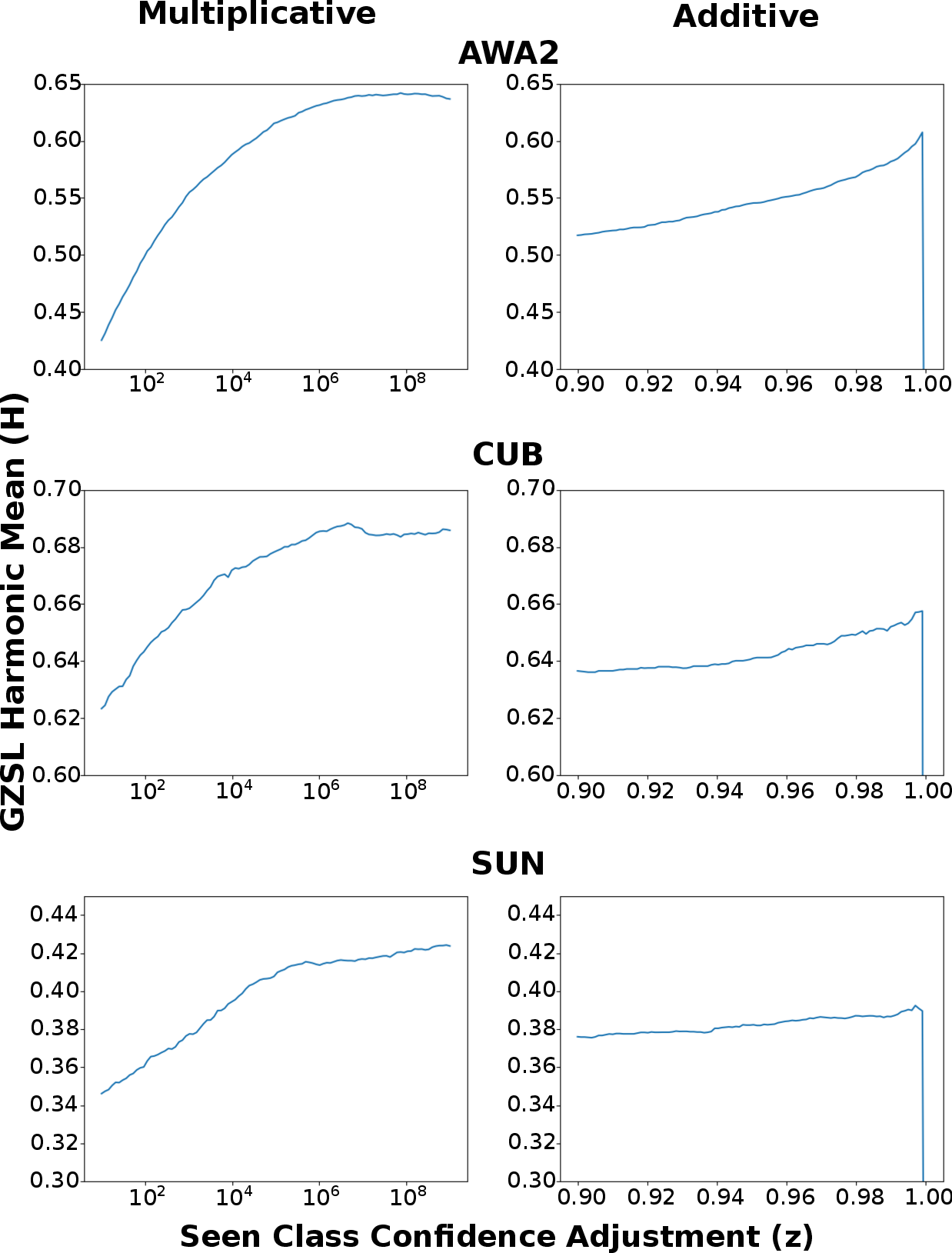}
\caption{An ablation study of the GZSL harmonic mean performance of DAZLE (with VINVL features) and RAJE when using addition and multiplication for calibrated stacking. The same vertical axes are used when plotting the multiplicative and additive performance in each dataset. Our proposed multiplicative approach for calibration exhibits greater performance over a larger portion of the graph while the previous additive approach has reduced performance and a sharp dip after its peak. Furthermore, additive calibration has the potential to sharply dip as it approaches 1 since it will begin classifying all examples as unseen.}
\label{fig:calibration}
\end{figure}
\section{Experiments}
\subsection{Experimental Setup}
We perform our experiments using the three widely adopted ZSL and GZSL benchmark datasets provided by \cite{ugly2018}. CUB \cite{cub2010} is a fine-grained bird species classification dataset with 150 seen and 50 unseen classes. With 312 human-annotated attributes, it is the most heavily annotated of the 3 benchmarked datasets and provides 7,057 training examples and 4,731 testing examples. SUN \cite{sun2012} is a scene classification dataset with 645 seen and 72 unseen classes. Only 10,320 training samples and 4,020 testing samples are provided along with 102 human-annotated attributes, meaning SUN has the least images per class of the 3 benchmarked datasets. AWA2 \cite{ugly2018} is an animal species classification dataset and is relatively coarse when compared to CUB and SUN. 23,527 training images, 13,795 test images, and 85 human-annotated attributes are provided. The traditional ZSL setting benchmark uses the top-1 accuracy (T1) performance on unseen class test set. The GZSL setting challenges models to classify both unseen and seen images in a single test set, such that unseen images may be misclassified as being from a seen category and vise versa. The harmonic mean (H) measures the trade-off between unseen and seen test set performance and serves as the primary metric for GZSL, with the unseen (u) and seen (s) accuracy included for additional transparency.
\par
In our implementations, visual features from 3 sources are tested: ResNet101 pre-trained on ImageNet \cite{ugly2018,resnet2016} providing the most least localized information, 2048x7x7 grid features (R=49) of ResNet101 fine-tuned using APN (model weights are only available for CUB) providing more localized information, and the top 30 (R=30) most confident proposals provided by VINVL \cite{vinvl2021} of dimension 2048 providing the most localized information. Combinations of the aforementioned visual features with 3 different base models are tested: SJE \cite{sje2015}, DAZLE \cite{dazle2020}, and our proposed PPN. Label attributes are sourced from human annotations provided by \cite{ugly2018}. As done in DAZLE \cite{dazle2020}, we semantically embed the human annotated attributes using word2vec \cite{w2v2013}. The optimization procedure for PPN utilizes the Adam optimizer with a learning rate of 0.001 following the procedure from \cite{vgse2022}. The proposed hyperparameters obtain their highest validation set performance at 0.1 across all tested benchmarks. Thus, all tested base models are out-of-the-box, meaning no dataset-specific hyperparameter settings are utilized.
\subsection{(Generalized) Zero-Shot Learning}
\label{sec:gzsl}
Table \ref{tab:ZSL_acc} explores the relationship between different visual feature extractors and specialized base models. With the use of VINVL features, DAZLE becomes a more competitive method for both the CUB and SUN dataset, while a slight decrease in performance occurs in AWA2. This may be caused in part by similarities between the categories in AWA2 and the ImageNet dataset, on which the ResNet101 feature extraction is pretrained. Consistent improvements in the performance of PPN can be observed as more features with more localized information are utilized. The use of both regularizers also consistently improves the performance of PPN with the exception of the SUN dataset, where only the $\mathcal{L}_{vis}$ regularizer contributes to improved performance The multiplicative corrections used to compensate for the large discrepancy between seen and unseen confidence for GZSL can be seen in Figure \ref{fig:calibration}. The plots show significant improvements in performance when using the multiplicative correction compared to the previously proposed additive correction. 

%% file: sec/7_conclusion.tex
\section{Conclusion and Future Work}
We propose a novel approach for localization in GZSL using region proposals from a pre-trained VL detector (VINVL) and utilize the provided proposals in to create part prototype which extract relevant information from attributes for each of these regions. Our ablation and analysis show that VINVL is a highly effective visual information source for GZSL and that our proposed Part Prototype Network can potentially serve as an improved foundation for future GZSL works. One potential avenue not explored in this work is enhancements of the visual features provided by VINVL, either through regularization like in generative approaches or additional training tailored to the zero-shot tasks.

\noindent {\bf Acknowledgments: }
The authors acknowledge support from the NSF RI Project VR-K \#1750082, CPS Project \#2038666. Any opinions, findings, and conclusions in this publication are those of the authors and do not necessarily reflect the view of the funding agencies. 